%% file: neurips_2023.tex
\documentclass{article}

\usepackage[most]{tcolorbox}
\PassOptionsToPackage{numbers, compress}{natbib}

\usepackage[final]{neurips_2024}

\include{math_commands}

\usepackage[utf8]{inputenc} %
\usepackage[T1]{fontenc}    %
\usepackage{hyperref}       %
\usepackage{url}            %
\usepackage{
booktabs}       %
\usepackage{amsfonts}       %
\usepackage{nicefrac}       %
\usepackage{microtype}      %
\usepackage{graphicx} 
\usepackage{multirow} 
\usepackage{amsmath}
\usepackage{amssymb}
\usepackage{mathtools}

\usepackage{booktabs}
\usepackage{listings}
\usepackage{xcolor}

\definecolor{codegreen}{rgb}{0,0.6,0}
\definecolor{codegray}{rgb}{0.5,0.5,0.5}
\definecolor{codepurple}{rgb}{0.58,0,0.82}
\definecolor{backcolour}{rgb}{0.95,0.95,0.92}

\lstdefinestyle{mystyle}{
    backgroundcolor=\color{backcolour},   
    commentstyle=\color{codegreen},
    keywordstyle=\color{magenta},
    numberstyle=\tiny\color{codegray},
    stringstyle=\color{codepurple},
    basicstyle=\footnotesize\ttfamily,      
    breaklines=true,                 
    captionpos=b,                    
    keepspaces=true,                 
    numbers=none,                              
    tabsize=2
}

\lstdefinelanguage{JSON}{
    basicstyle=\normalfont\ttfamily,
    numbers=none,
    breaklines=true,
    frame=lines,
    backgroundcolor=\color{backcolour},
    stringstyle=\color{codepurple},
    keywordstyle=\color{blue},
    morekeywords={null,true,false}
}

\title{Conv-CoA: Improving Open-domain Question Answering in Large Language Models via Conversational Chain-of-Action}

\author{
  Zhenyu Pan \\ Northwestern University \\ \texttt{zhenyupan@u.northwestern.edu} \And  Haozheng Luo \\ Northwestern University \\ \texttt{RobinLuo2022@u.northwestern.edu}  \And Manling Li \\ Northwestern University \\ \texttt{manling.li@northwestern.edu} \And Han Liu \\ Northwestern University \\ \texttt{hanliu@northwestern.edu}
  }

\begin{document}

\maketitle

\begin{abstract}

We present a Conversational Chain-of-Action (Conv-CoA) framework for Open-domain Conversational Question Answering (OCQA). Compared with literature, Conv-CoA addresses three major challenges: (i) unfaithful hallucination that is inconsistent with real-time or domain facts, (ii) weak reasoning performance in conversational scenarios, and (iii) unsatisfying performance in conversational information retrieval. Our key contribution is a dynamic reasoning-retrieval mechanism that extracts the intent of the question and decomposes it into a reasoning chain to be solved via systematic prompting, pre-designed actions, updating the Contextual Knowledge Set (CKS), and a novel Hopfield-based retriever. 
Methodologically, we propose a resource-efficiency Hopfield retriever to enhance the efficiency and accuracy of conversational information retrieval within our actions. Additionally, we propose a conversational-multi-reference faith score (Conv-MRFS) to verify and resolve conflicts between retrieved knowledge and answers in conversations. Empirically, we conduct comparisons between our framework and 23 state-of-the-art methods across five different research directions and two public benchmarks. These comparisons demonstrate that our Conv-CoA outperforms other methods in both the accuracy and efficiency dimensions.

\end{abstract}

\section{Introduction}

\input{1introduction}\label{sec:introduction}

\section{Related Work}
\input{2relatedWork}\label{sec:relatedWork}

\vspace{-0.2in}
\section{Methodology}
\input{3methdology}\label{sec:methodology}

\section{Experiments}

\input{
4experiments}\label{sec:experiments}

\section{Conclusions and Future Work}
\input{5summary}\label{sec:summary}
\newpage

\clearpage

\section*{Broader Impact}
Our research methodology enhances understanding and problem-solving across various domains, including AI research, by producing clearer and more comprehensible results. However, this method might oversimplify complex issues by breaking them down into discrete parts, potentially overlooking nuances and interrelated elements. Additionally, relying heavily on this approach could limit creative problem-solving, as it encourages a linear and structured process that may impede unconventional thinking.

\input{6reference}\label{sec:references}

\clearpage
\newpage
\setlength{\abovedisplayskip}{10pt}
\setlength{\abovedisplayshortskip}{10pt}
\setlength{\belowdisplayskip}{10pt}
\setlength{\belowdisplayshortskip}{10pt}
\appendix
\part*{Supplementary Material}
\input{appendix}
\end{document}

%% file: math_commands.tex
\usepackage[utf8]{inputenc}
\usepackage{amsmath}
\usepackage{amssymb}
\usepackage{mathtools}
\numberwithin{equation}{section}
\usepackage{slashed}
\usepackage{braket}
\usepackage{enumitem}
\usepackage[bottom]{footmisc}
\usepackage{cite}
\usepackage{graphicx}
\usepackage{bm}
\usepackage{subfig}
\usepackage{physics}
\usepackage{xcolor}
\usepackage{natbib}
\usepackage{mdframed}

\usepackage{CJK}

\usepackage[linesnumbered,ruled]{algorithm2e}
\SetKwComment{Comment}{/* }{ */}
\SetKwInput{kwInit}{Initialization}
\SetKwRepeat{Do}{do}{while}%

\renewenvironment{figure}[1][]{
  \begin{originalfigure}[#1]
    \begin{mdframed}[linecolor=black!0,backgroundcolor=black!1]
}{
    \end{mdframed}
  \end{originalfigure}
}

\def\0{{(0)}}
\def\1{{(1)}}
\def\2{{(2)}}

\def\<{\langle }
\def\>{\rangle }

\newcommand{\bea}{\begin{eqnarray}}
\newcommand{\eea}{\end{eqnarray}}

\newcommand{\Sparsemax}{\mathop{\rm{Sparsemax}}}

\usepackage{slashed}

\def\({\left(}
\def\){\right)}
\def\[{\left[}
\def\]{\right]}

\usepackage{dashbox}
\usepackage{xcolor}
\usepackage{colortbl}
\usepackage{arydshln}
\definecolor{lightyellow}{rgb}{1.0, 0.95, 0.7}
\definecolor{Blue}{rgb}{0.0, 0.4, 1.0}
\definecolor{blue}{rgb}{0,0,1}
\definecolor{darkgreen}{rgb}{0.,0.6,0.}

\definecolor{colorA}{rgb}{1,0,0}
\definecolor{colorB}{rgb}{0,0.3,1}
\definecolor{colorC}{rgb}{0.9,0.8,0.2}
\definecolor{colorD}{rgb}{0,0.65,0}
\definecolor{lesslightgray}{rgb}{0.5,0.5,0.5}

\definecolor{light-gray}{gray}{0.95}

\newcommand{\calT}{\mathcal{T}}

\newcommand{\bI}{\bm{I}}

\newcommand{\bK}{\bm{K}}

\newcommand{\bQ}{\bm{Q}}
\newcommand{\bR}{\bm{R}}

\newcommand{\bV}{\bm{V}}
\newcommand{\bW}{\bm{W}}

\newcommand{\bY}{\bm{Y}}
\newcommand{\bZ}{\bm{Z}}

\newcommand{\by}{\bm{y}}

\newcommand{\sT}{ \mathsf{T} }

\let\cite\citep

\usepackage{amsthm}
\usepackage{nicefrac}

\theoremstyle{definition}

\numberwithin{equation}{section}
\numberwithin{theorem}{section}

\usepackage{lipsum}

\makeatletter
\let\save@mathaccent\mathaccent
\newcommand*\if@single[3]{%
    \setbox0\hbox{${\mathaccent"0362{#1}}^H$}%
    \setbox2\hbox{${\mathaccent"0362{\kern0pt#1}}^H$}%
    \ifdim\ht0=\ht2 #3\else #2\fi
}
\newcommand*\rel@kern[1]{\kern#1\dimexpr\macc@kerna}
\newcommand*\widebar[1]{\@ifnextchar^{{\wide@bar{#1}{0}}}{\wide@bar{#1}{1}}}
\newcommand*\wide@bar[2]{\if@single{#1}{\wide@bar@{#1}{#2}{1}}{\wide@bar@{#1}{#2}{2}}}
\newcommand*\wide@bar@[3]{%
    \begingroup
    \def\mathaccent##1##2{%
        \let\mathaccent\save@mathaccent
        \if#32 \let\macc@nucleus\first@char \fi
        \setbox\z@\hbox{$\macc@style{\macc@nucleus}_{}$}%
        \setbox\tw@\hbox{$\macc@style{\macc@nucleus}{}_{}$}%
        \dimen@\wd\tw@
        \advance\dimen@-\wd\z@
        \divide\dimen@ 3
        \@tempdima\wd\tw@
        \advance\@tempdima-\scriptspace
        \divide\@tempdima 10
        \advance\dimen@-\@tempdima
        \ifdim\dimen@>\z@ \dimen@0pt\fi
        \rel@kern{0.6}\kern-\dimen@
        \if#31
        \overline{\rel@kern{-0.6}\kern\dimen@\macc@nucleus\rel@kern{0.4}\kern\dimen@}%
        \advance\dimen@0.4\dimexpr\macc@kerna
        \let\final@kern#2%
        \ifdim\dimen@<\z@ \let\final@kern1\fi
        \if\final@kern1 \kern-\dimen@\fi
        \else
        \overline{\rel@kern{-0.6}\kern\dimen@#1}%
        \fi
    }%
    \macc@depth\@ne
    \let\math@bgroup\@empty \let\math@egroup\macc@set@skewchar
    \mathsurround\z@ \frozen@everymath{\mathgroup\macc@group\relax}%
    \macc@set@skewchar\relax
    \let\mathaccentV\macc@nested@a
    \if#31
    \macc@nested@a\relax111{#1}%
    \else
    \def\gobble@till@marker##1\endmarker{}%
    \futurelet\first@char\gobble@till@marker#1\endmarker
    \ifcat\noexpand\first@char A\else
    \def\first@char{}%
    \fi
    \macc@nested@a\relax111{\first@char}%
    \fi
    \endgroup
    }
\makeatother

%% file: 1introduction.tex
This work proposes a conversational reasoning-retrieval framework to enhance both the efficiency and quality of Open-domain Conversational Question Answering (OCQA), tailored to surpass the architecture of traditional Retrieval Augmented Generation (RAG) methods.
This work addresses three major challenges in applying RAG to answer conversational questions: (i) \textbf{weak reasoning}, where large language models (LLMs) struggle to acquire information from specific heterogeneous sources, (ii) \textbf{unfaithful hallucinations}, where the response may not align with real-time or domain-specific facts, and (iii) \textbf{unsatisfying retrieval}, where the traditional dense information retriever (IR) cannot get the intent of questions and fails in the conversational scenario.

To enhance the reasoning, faithfulness, and conversational IR, previous approaches such as chain-of-thought-based work \cite{saparov2022language, yao2023tree} prompt LLMs to answer complex questions step by step. The other work proposes RAG-based prompting frameworks such as agents \cite{pan2024chainofaction}. However, they aim to solve single-round questions and fail in complex conversations. More recent work focuses on the retrieval phase and aims to improve the query quality and retriever capability. While CONQRR and ReExCQ \cite{wu2021conqrr,mo2023learning} are the representative query-reformulation methods to rewrite and expand the current query with historical conversations, they need extra pre-trained model with much training expense. We argue that training a model for query reformulation is unnecessary; instead, prompting methods can harness the existing LLMs to serve this purpose. ConvSDG and CONVAUG \cite{mo2024convsdg,chen2024generalizing} try to enhance the conversational retriever by generating different levels of conversations. However, their performance still cannot be beyond the theoretical upper bound of traditional dense retrievers, and their consumption is still expensive. In summary, the key challenge of current OCQA lies in designing a framework tailored for conversational QA scenarios that integrates prompting with RAG and devises an innovative conversational retriever that surpasses traditional retriever architecture in efficacy.

\begin{figure*}[t]
    \centering
    \includegraphics[width=1\linewidth]{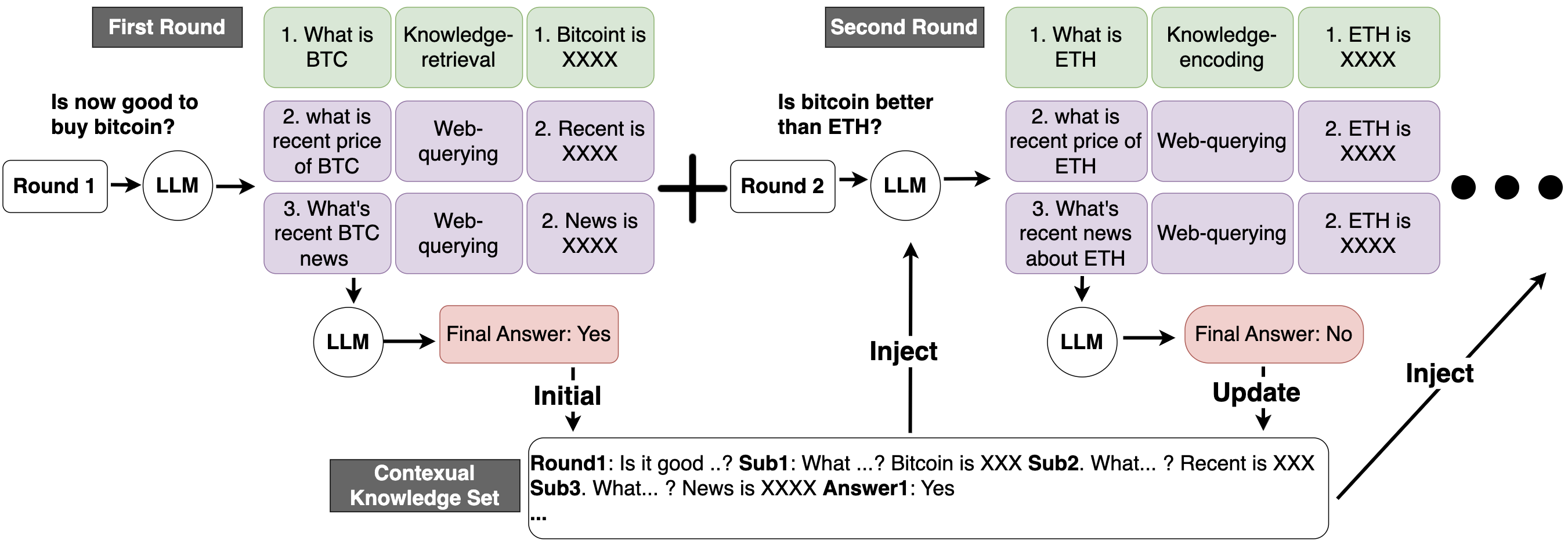}
    \vspace{-0.2in}
    \caption{Overview of Conv-CoA. It starts by injecting the initial question into a prompt, creating an action chain (AC) via the LLM. Each node in the AC represents a sub-question and an initial answer. Specific actions verify if the initial answer needs modification based on retrieved data. If the confidence in the initial answer is lower than in the retrieved data, an action prompts a change. The AC contents are stored in a contextual knowledge set (CKS) and updated at the end of each turn. In future turns, the CKS and the current question are combined to regenerate the AC, using different prompt templates to generate sub-questions for missing content, thus reducing information overlap.}
    \label{fig:framework}
    \vspace{-0.25in}
\end{figure*}

To this end, we propose \textbf{Conv-CoA} framework, to deliver an accurate, faithful, and fast OCQA. In Conv-CoA, we design two actions: web-querying and info-searching. The former could search for the newest information from the Internet, while the latter could retrieve knowledge from storage (i.e., specific domain documents). Because we also need to rank the search results in the web-querying action, data retrieval is the most essential part of both actions. Inspired by the Modern Hopfield models showcasing fast convergence and exponential memory capacity in various domains \cite{schimunek2023contextenriched,widrich2020modern}, we design a resource-efficiency Hopfield Retriever, tailored to surpass the traditional dense retriever architecture in searching with enhanced speed and precision.
Meanwhile, we design a systematic prompting method to decompose a complex question into a reasoning chain with many subquestions. As shown in Figure~\ref{fig:framework}, our method injects the first turn question into a designed prompt and then constructs an action chain (AC) by the LLM. Each node in the AC represents a sub-question and an initial answer generated by LLM. Then, we assign a specific action on each node to verify if the initial answer needs to change with the retrieved data. If the confidence of the initial answer is lower than that of retrieved data, the action requests to change. Finally, we store the contents of AC in a contextual knowledge set (CKS) and update it at each end of a future turn. At the same time, the LLM generates the final answer based on the updated CKS. In the following turns, we iterative combine the CKS and the current turn's question to repeat the AC generation. In particular, the framework uses a different prompt template for it. We design a rule to generate the subquestions that we do not have relevant content yet based on the newest CKS. Therefore, it helps to decrease the overlap of the retrieved information by updating the CKS at the end of future turns. 

In summary, Conv-CoA is the first work that enables a faithful, accurate, and fast OCQA by incorporating the prompting method and improved Hopfield retriever. Our main contributions are as follows:
\vspace{-0.2in}
\begin{itemize}
    \item We propose a Conv-CoA framework that could decompose each question into many sub-questions to resolve by Hopfield-enhanced actions with verification by Conv-MRFS score. 
    \item We propose a resource-efficiency Hopfield retriever to solve the limitations of traditional dense retrievers and speed up the retrieval, decreasing latency with no sacrifice in accuracy.
    \item We design a systematic prompting to construct the Action Chain of each turn based on the updating CKS, enabling less overlap of the retrieval to decrease the latency further.
    \item Experimental results demonstrate that our framework can outperform other methods in public benchmarks with a high degree of accuracy and efficiency. 
\end{itemize}

%% file: 2relatedWork.tex
To enhance performance in OCQA, recent work mainly focuses on Retrieval Augmented Generation (RAG) methods. We categorize these methods into two distinct phases in alignment with the RAG life cycle:
(1) Retrieval Phase: methods aim to enhance query quality and retrieval efficacy, facilitating the acquisition of more pertinent information to support LLM in generating answers. 
(2) Generation Phase: methods aim to propose prompting methods to aid LLM in conducting reasoning processes, thereby enabling more rational and precise answer generation.

Hence, we first introduce the work in the retrieval phase from two perspectives: query quality and retriever capability. Then, we explore various prompting methods designed to synergize with retrieval phase. In addition, we also introduce some hopfield-based methods that motivate us to propose our memory-enhanced conversational retriever. 

\subsection{Retrieval Phase}
\subsubsection{Query Reformulation Methods}
Conversational Query Reformulation (CQR) uses query rewriting and query expansion based on a conversational context to improve conversational retrieval performance. Compared with other conversational retrieval methods, CQR could directly reformulate the original conversation-based query into a standalone query as input to off-the-shelf retrievers without fine-tuning. 
While previous CQR work addresses conversational retrieval by using human-rewritten queries or querying expansion methods \citep{lin2021contextualized,lin2020conversational, mo2023convgqr}, they always get a sub-optimal and require a separate model trained by collecting lots of human rewrites.

To address these drawbacks, CONQRR \citep{wu2021conqrr} directly optimizes the query rewriting model to the retrieval. More recently, IterCQR and IQR (\cite{jang2023itercqr,ye2023enhancing}) conduct query reformulation without relying on human rewrites by prompting the large language models (LLMs). ReExCQ (\citep{mo2023learning}) focuses on expanding the current query with selected relevant historical queries.
However, they need a large storage capacity and training time when generating candidates during the training process.

\subsubsection{Enhanced Retrieval Methods}
Conversational Retrievers train previous information retrievers on conversational datasets using more complicated training strategies and loss functions. It aims to enhance the retriever's ability to search for relevant information within conversational situations.
LeCoRE \citep{mao2023learning} considering knowledge distillation, InstructorR \citep{jin2023instructor} utilizing LLMs to predict the relevance score between the session and passages, and SDRConv \citep{kim2022saving} that includes mining additional hard negatives. In addition, ConvSDG and CONVAUG \citep{mo2024convsdg,chen2024generalizing} are the most recent work about utilizing LLM to generate conversations. ConvSDG explores the dialogue/session-level and query-level data generation separately. CONVAUG generates multi-level augmented conversations to capture the diverse nature of conversational contexts. HAConvDR \citep{mo2024history} incorporating context-denoised query reformulation and automatic mining of supervision signals based on the actual impact of historical turns. 

However, all of them focus on augmenting the training data for conversational dense retrievers. Despite these advancements, the performance upper limit of these methods still cannot break through the theoretical upper limit of conversational dense retrievers.

\subsection{Prompting Methods}

Prompting methods aim to direct the LLMs to follow given instructions. The commonly used method of few-shot prompting \cite{kaplan2020scaling} facilitates in-context learning that guides LLMs to comply with directives and respond to queries using just a few examples. Methods like Chain-of-Thought (CoT) \cite{NEURIPS2022_9d560961} and its enhanced versions \cite{wang2022self,saparov2022language} seek to steer LLMs towards breaking down intricate tasks into logical sequences of reasoning, thereby improving performance. The Chain-of-Action (CoA) \cite{pan2024chainofaction} integrates the reasoning capabilities of CoT with the information retrieval prowess of external retrievers, crafting a collaborative design that culminates in a faithful and multimodal QA system. However, it lacks support for OCQA and does not overcome the limitations of traditional dense retrieval methods. 

\subsection{Modern Hopfield Models and Hopfield Memory}

Classical Hopfield models \citep{hopfield1984neurons,hopfield1982neural,krotov2016dense} are pivotal in emulating the human brain's associative memory, emphasizing memory pattern storage and retrieval. The resurgence in Hopfield model research is driven by enhanced memory storage capacities \citep{krotov2016dense, demircigil2017model}, innovative architectural designs \citep{wu2024uniform,wu2023stanhop,hoover2023energy,seidl2022improving,furst2022cloob,ramsauer2020hopfield}, and their biological plausibility \citep{kozachkov2022building,krotov2020large}. Modern Hopfield models \citep{hu2024nonparametric,hu2024computational,wu2023stanhop,hu2023SparseHopfield,hopfeildblog2021,ramsauer2020hopfield} showcase fast convergence and exponential memory capacity, linking them to Transformer architecture as advanced attention mechanisms. Their application spans various domains, including drug discovery \citep{schimunek2023contextenriched}, immunology \citep{widrich2020modern}, and time series forecasting \citep{wu2023stanhop,hu2023SparseHopfield,auer2023conformal}, signifying their influence on future large-scale model designs. Our research focuses on leveraging the rapid convergence and vast memory capacity of modern Hopfield models to efficiently retrieve knowledge from memory spaces. This approach aims to minimize latency in question-answer interactions within our Conversational CoA framework.

%% file: 3methdology.tex
\subsection{Task Definition}
Our framework aims to generate answers aligned with the current conversational question, denoted by $q_n$ for the $n$-th turn. This QA process is historically contextual, leveraging previous dialogue turns denoted by $\mathcal{H} = \{q_i\}_{i=1}^{n-1}$. It is essential for the framework to optimize the formulation of each question $q_i$ to accurately capture the user's intended query content $\mathcal{O}_n$. Utilizing the reasoning capabilities of large language models (LLMs), the framework decomposes the optimized question into a chain of  $k$ sub-questions  $\{\mathcal{R}_{n1}, \mathcal{R}_{n2}, \ldots, \mathcal{R}_{nk}\}$, each aimed at a specific aspect of the main query. For each sub-question $\mathcal{R}_{ni}$, the system retrieves the most relevant information passage $\mathcal{I}_{ni}$ from a corpus of external data sources, culminating in a set $\{\mathcal{I}_{ni}\}_{i=1}^k$ that aids in generating the final answer $a_n$. Thus, the abilities to optimize questioning $\mathcal{O}$, to chain reasoning $\mathcal{R}$, and to retrieve pertinent information $\mathcal{I}$ are pivotal. The goal of this paper is to propose a framework that integrates disparate external information sources, empowering the LLM to dissect queries effectively and retrieve related content swiftly, leading to the provision of the most precise answer $\mathcal{A}_n$.

\vspace{-0.2in}
\subsection{Overview}

As shown in Figure~\ref{fig:framework}, we utilize in-context learning to inject the initial user query into a structured prompt template, generating an Action Chain (AC) via a LLM. Each node within the AC includes a sub-question and a preliminary response, with specific actions executed to adjust these responses based on the confidence comparison between the initial answers and newly retrieved data. Concurrently, we maintain and iteratively update a Contextual Knowledge Set (CKS) at each conversation turn, which stores essential information and reduces redundancy in retrieved data. This dynamic updating supports the decomposition of subsequent queries into sub-questions using an adaptive prompt template, enhanced by our Hopfield-based retriever, which ensures rapid and accurate information retrieval, enriching the contextual understanding and response precision throughout the conversation.

\vspace{-0.2in}

\subsection{Contextual Knowledge Set}
The Contextual Knowledge Set (CKS) is a structured data format designed to store critical information from each round of a conversation. CKS is in JSON format, which facilitates easy integration and manipulation in data-driven applications. Each entry in the CKS records various components of the dialogue, including the original question posed during the round, the optimized question that refines or extends the original inquiry, detailed sub-questions that break down the main question into more reasonable parts, summarized information that has been retrieved in response to each sub-question, and the final answer concluded at the end of the discussion. This structure not only preserves the flow of conversation but also enriches the contextual understanding of each interaction. Below is an example of how CKS organizes and presents this information:
\noindent\begin{minipage}{\linewidth}
\begin{lstlisting}[language=JSON, caption=Example of a Contextual Knowledge Set JSON Structure]
{
  "Contextual knowledge set": [
    {
      "round": 1,
      "original_question": "What is the process of ...?",
      "optimized_question": "Explain the steps involved ...",
      "sub_questions": {
        "sub1": "What are the light-dependent reactions ...",
        "sub2": "What are the light-independent reactions ...",
        "sub3": "How do plants convert sunlight into ..."
      },
      "information_summaries": {
        "infor1": "Light-dependent reactions use light ...",
        "infor2": "Light-independent reactions, or the ...",
        "infor3": "Plants convert sunlight into chemical ..."
      },
      "answer": "Photosynthesis is a process where..."
    },
    ...
  ]
}
\end{lstlisting}
\end{minipage}
\vspace{-0.2in}
\subsection{Action chain generation} 
We have two different action chain generation stages, the initial and normal stages. 

\textbf{Initial stage:} 
For the initial stage, we inject the first-round question and descriptions of actions into a prompting template. Then, we prompt LLM to decompose the question into sub-questions with guess answers. The prompting template is shown in \ref{prompt:initial}. After that, the actions retrieve the related information and correct these sub-questions. Finally, LLM generates the final answer from the processed action chain. In addition, the framework updates the CKS by the processed action chain.

\textbf{Normal stage:} 
In the normal stage, we design a prompt to let LLM extract the N-round intent question from the updated CKS and original N-round question, then decompose it into sub-questions. After that, actions do the retrieval and answer correction again. Finally, LLM generates the final N-round answer and updates the process into the CKS. The prompting template is shown in \ref{prompt:initial}:

\subsection{Action Implementation} 
\subsubsection{Action design}
\paragraph{Web-querying.}
Our web-querying action leverages search engines such as Google Search, employing a query strategy to retrieve relevant Internet content. Initially, the action searches for keywords from the specified sub-question $\mathcal{R}_{nk}$, generating a list of results. If the "Missing\_flag" is set to "True", we select the top-k results and extract the content directly from their web pages. Conversely, if the flag is "False", we collate the titles $T$ and snippets $Sn$ from the top M pages. Each title and snippet pair ${\{T_m, Sn_m\}}$ is then transformed into a 1536-dimension vector $Emb\{T_m | Sn_m
\}$ using the OpenAI's text-embedding-ada-002 model \cite{openai2023gpt4}. We perform a similar vector transformation for the sub-question and its guess answer $\{\mathcal{R}_{nk}, \mathcal{G}_{nk}\}$. Subsequently, we compute the similarity between each vector $Emb\{T_m | Sn_m\}$ and $Emb\{\mathcal{R}_{nk} | 
\mathcal{G}_{nk}\}$, filtering out pages with similarity scores below 0.8. The contents from the pages with high similarity are then extracted and re-evaluated to rank and select the top-k final pages, from which we retrieve the ultimate information required for the action.

\paragraph{Knowledge-retrieval.}

In our framework, documents sourced from diverse platforms, including Wikipedia, are pre-processed using a encoder designed to transform textual content into embedding vectors. Each document is segmented into multiple chunks determined by their length, following which these segments are encoded into vectors. These vectors are subsequently cataloged in a vector database, indexed for efficient retrieval. For the retrieval process, we leverage our novel Hopfield retriever, which utilizes these pre-encoded vectors to respond to queries effectively, ultimately yielding the top-k most relevant results based on similarity metrics. The details of the encoding and retrieval mechanisms will be further elaborated in the subsequent section of this paper \ref{sec:hopfield_retriever}.

\vspace{-0.2in}
\begin{table}[htp]
\centering
\caption{\textbf{Retrieval Performance on Conversation-Based Datasets} We conduct experiments on conversation-based datasets, testing 12 baselines in a conversational retrieval task using Mean Reciprocal Rank (MRR) and Recall@10 scores. We report average MRR and Recall@10 metrics, omitting variance as all values are $\leq$ 2\%. The best results are highlighted in bold, while the second-best are underlined. Across most configurations, the Hopfield Retrieval (HR) with training outperforms all baselines, and HR without retrieval surpasses several of the retrieval methods. }
\resizebox{0.85\textwidth}{!}{%
\begin{tabular}{cccccc}
\toprule
\multirow{2}{*}{Category} & \multirow{2}{*}{Model}   & \multicolumn{2}{c}{QReCC}                    & \multicolumn{2}{c}{TopiOCQA}         \\ \cline{3-6} 
                          &                          & MRR        & Recall@10    & MRR           & Recall@10  \\ \midrule
\multirow{5}{*}{QR}    & T5QR                     & 23.0       & 37.6      & 34.5          & 53.1    \\
                          & CONQRR                   & 24.3          & 39.8         & 41.8          & \underline{65.1}    \\
                          & ConvGQR                  & 25.6       & 41.8      & 42.0          & 63.5    \\
                          & IterCQR                  & 42.9       & 65.5      & 26.3          & 42.6    \\
                          & IQR                      & 45.0       & 67.3      & 31.3             & 49.8       \\  \midrule
\multirow{3}{*}{RE}       & ReExCQ                   & 18.5       & 28.9      & 10.8          & 24.1    \\
                          & ConvSDG                  & 39.8          & 43.7         & 21.4          & 37.8    \\
                          & CONVAUG                  & 52.7       & 75.6      & 35.0          & 57.9    \\ \midrule
                          
\multirow{4}{*}{RB}       & HAConvDR                 & 48.5       & 72.4      & 30.1          & 50.8    \\                                              
                          & LeCoRE                   & 51.1       & 73.9      & 32.0          & 54.3    \\
                          & InstructorR              & 52.9       & \underline{77.7}      & \underline{38.5}          & 62.1    \\
                          & SDRConv                  & \underline{53.0}       & 76.1      & -             & -       \\ \midrule
\multirow{2}{*}{\textbf{HR}} & HR w/o Training      & 45.1$^\dag$ & 70.2$^\dag$ & 33.7$^\dag$  & 59.2$^\dag$  \\ 
                          & HR w/ Training           & \textbf{68.7$^\dag$} & \textbf{83.5$^\dag$} & \textbf{60.3$^\dag$}  & \textbf{74.8$^\dag$}  \\
\bottomrule
\end{tabular}
}
\label{tab:result}
\end{table}
\vspace{-0.1in}

\subsubsection{Resource-Efficiency Hopfield Retriever}
\label{sec:hopfield_retriever}

Based on the existing Hopfield model \citep{xu2024bishop, hu2023SparseHopfield, hopfeildblog2021}, we propose a resource-
efficiency Hopfield memory retrieval model designed to extract top-$k$ relevant information and solution paths from existing conversation turns and the knowledge base. This model aims to enhance the efficiency and accuracy of information retrieval by leveraging historical interactions and structured data sources. 

\paragraph{Encoder.}
In this research, we employ separate BERT-based \citep{devlin2019bert} networks for the question and passage encoders, using the [CLS] token representation as the output, with each representation being 768-dimensional. To enhance efficiency, we utilize an 8-bit quantized version \cite{dettmers2022gpt3} of the BERT\_based model. Previous studies \citep{bondarenko2024quantizable} have shown that large foundation models often suffer from numerous outliers that degrade efficiency and quantization performance due to attention-weight explosions caused by these outliers. In response, we adopt the $\mathtt{OutEffHop}$ \citep{hu2024outlierefficient} variant of BERT, which mitigates performance loss associated with these challenges.

\paragraph{Interface.} During inference, we implement a speed-up strategy to enhance the adaptability of the Hopfield model in large-scale scenarios, achieving high speed with minimal impact on accuracy. As shown in \ref{sec:SparseHopfield}, given the extensive time required for querying a large-scale knowledge database, we adapt the $\mathtt{SparseHopfield}$ model to segment the memory pattern into several parts:
\begin{align}
    \bY &= \{\by_1, \by_2, \by_3, \cdots, \by_n\} \\
    \bZ_i &=\mathtt{SparseHopfield}(\bR,\by_i) = \Sparsemax(\beta  {\bR}\bW_Q \bW_K^\sT\by_i^\sT)\by_i\bW_K \bW_V.
    \label{eq:1}
\end{align}

This division allows for parallel processing, which does not degrade the performance of memory retrieval. Consequently, we evenly divide our Hopfield memory pattern into $k$ parts, with each segment independently retrieving the most suitable memory pattern as its output.

\paragraph{Hopfield-based Retrieval.} 

In our Hopfield-based memory retrieval model, we transition from retrieving standard memory patterns to directly extracting knowledge from conversations. It is crucial to train the model to refine the retrieval dynamics $\calT$, ensuring accurate convergence on relevant question-passage pairs. Following the characteristics of Hopfield retrieval, we offer a training version and an interference-direct mode that retrieves and ranks the top-k relevant passages, enabling the generation of answers, akin to the approach described by \cite{davydov2023retrieving}. In the training version, we adopt the loss function from the DPR model described by \cite{karpukhin2020dense} to train our Hopfield-based retrieval system.

Let $D = \{q_i, y_{i,1}^{+}, y_{i,1}^{-}, y_{i,2}^{-}, \dots, y_{i,k}^{-}\}_{i=1}^m$ represent the training data, which comprises $m$ instances. Each instance includes one question $q_i$, one relevant (positive) memory sets $y_{i,j}^{+}$, and with $k$ irrelevant (negative) memory sets $y_{i,j}^{-}$. positive passages are paired with different questions from the training set to form negative pairs for retrieval. This method of using gold passages from other questions as negatives enhances computational efficiency and yields high performance.

\subsubsection{Answering Verification}
To verify conflicts between generated answers and retrieved information in conversations, we introduce the Conversational-Multi-Reference Faith Score (Conv-MRFS). This metric evaluates the consistency of generated answers with the conversation history (CKS).

\textbf{Overview of Conv-MRFS}: 
The Conv-MRFS framework involves extracting relevant segments from the conversation history to serve as references. The core of Conv-MRFS is the faith score, which measures alignment between the generated answer and each reference segment based on precision, recall, and average word length.

\textbf{Components of the Faith Score}: 
The faith score $S$ is a composite metric incorporating three components: Precision (P), Recall (Rcl), and Average Word Length (AWL). These components are weighted to reflect their importance in the evaluation process: $\text{S} = \alpha \times P + \beta \times Rcl + \gamma \times AWL.$
 Here, $\alpha, \beta, \gamma$ are weights for precision, recall, and average word length, summing to 1 for normalization.

\textbf{Precision and Recall:}
Precision (P) quantifies the fraction of relevant instances within the generated answer consistent with the conversation history and 
Recall (Rcl) measures the fraction of relevant instances from the conversation history captured by the generated answer:

\begin{equation}
P = \frac{\text{number of consistent items}}{\text{total number of items in } A_n}, Rcl = \frac{\text{number of consistent items}}{\text{total number of relevant items in } CH}
\end{equation}

\textbf{Average Word Length:}
Average Word Length (AWL) represents the mean length of words in the generated answer, indicating verbosity and informativeness:

\begin{equation}
AWL = \frac{\text{sum of lengths of all words in } A_n}{\text{total number of words in } A_n}
\end{equation}

\textbf{Faith Score Calculation:}
For each segment $CH_i$ in the conversation, we compute the faith score $S(CH_i, A_n)$. The Conv-MRFS is the maximum faith score across all segments.

\textbf{Threshold Decision:}
We set a threshold $T$ for answer. If the Conv-MRFS exceeds $T$, the answer is considered faithful. Otherwise, the answer is revised to better align with the retrieved information.

%% file: 4experiments.tex
\begin{table}[t]
\centering
\caption{\textbf{Perfromance of Question Answering Abilities } We conduct an experiment on question answering using 11 baselines of chain-of-thought methods, evaluated with the GPT Exact Match Score (GPT-EM). We report the average GPT-EM score, omitting variance since all values are $\leq$ 2\%. The best results are highlighted in bold, while the second-best are underlined. Across most configurations, Conv-CoA achieves the best performance on the question-answering tasks in the TopiOCQA (Topi) and QReCC datasets.}
\resizebox{\textwidth}{!}{%
\begin{tabular}{cccccccccccccc}
\toprule
  & $0$-shot & Few & CoT & SC & ToT & LM 
  & ToolF & SA & React  & DSP & CoA 
 & Ours \\
\midrule
QReCC & 18.4 & 18.4 & 30.6 & 67.4 & 20.4 & 22.4 & 50.0 & 34.5 & 37.3 & 38.1  & \underline{69.7} & \textbf{71.2}\\
Topi & 23.2 & 28.2 & 35.4 & \underline{78.8} & 22.4 & 48.0 & 30.7 & 51.0 & 40.8 & 28.6  & 56.9 & \textbf{83.7}\\
\bottomrule
\end{tabular}
}
\label{tab:my_label}
\vspace{-0.2in}
\end{table}

In this section, we first compare our Conversational CoA framework with recent state-of-the-art baselines across various public benchmarks, followed by an in-depth analysis of three aspects: effectiveness of retrieval, efficiency of retrieval, and rate of misleading information.
Note that \textbf{GPT-3.5-Turbo} serves as the LLM in all LLM-based methods and as the reader for all retrievers.

\paragraph{Datasets}
We select two public benchmarks. TopiOCQA \cite{adlakha2022topiocqa}, an open-domain conversational dataset with topic switches on Wikipedia, contains 3920 conversations with information-seeking questions and free-form answers. On average, a conversation in TopiOCQA spans 13 question-answer turns and involves four topics (documents). QReCC (Question Rewriting in Conversational Context) \cite{anantha2020open}, an end-to-end open-domain question-answering dataset comprising 14K conversations with 81K question-answer pairs.

\paragraph{Baselines}
We compare our framework with three types of state-of-the-art approaches. 

1. Query-reformulation-based approaches:

\textit{Query Rewriting (\textbf{QR})}: T5QR \citep{lin2020conversational}, CONQRR \citep{wu2021conqrr}, ConvGQR \citep{mo2023convgqr}, IterCQR \citep{jang2023itercqr} and IQR \citep{ye2023enhancing}.

\textit{Query Expansion (\textbf{RE})}: 
ReExCQ \citep{mo2023learning}, ConvSDG \citep{mo2024convsdg}, and CONVAUG \citep{chen2024generalizing}

2. Ad-hoc search dense retriever-based approaches (\textbf{RB})

HAConvDR \citep{mo2024history} incorporating context-denoised query reformulation and automatic mining of supervision signals based on the actual impact of historical turns, LeCoRE \citep{mao2023learning} considering knowledge distillation, InstructorR \citep{jin2023instructor} utilizing LLMs to predict the relevance score between the session and passages, and SDRConv \citep{kim2022saving} that includes mining additional hard negatives. 

3. Prompting-based approaches

\textit{Prompting only (\textbf{PO})}: Zero-shot Prompting,
Few-shot Prompting (Few), Chain-of-Thought (CoT) \cite{NEURIPS2022_9d560961}, Self Consistency (SC) \cite{wang2022self}, Tree of Thought (ToT) \cite{yao2023tree}, and Least-to-Most (LM) \cite{zhou2022least}

\textit{Prompting with RAG (\textbf{PRAG})}:
ToolFormer (ToolF) \cite{Schick2023ToolformerLM},Self-Ask (SA) \cite{press2022measuring}, React \cite{yao2023react}, DSP \cite{
khattab2022demonstrate}, and CoA \cite{pan2024chainofaction}.

4. Ablation study about Hopfield-based baselines

\textit{Hopfield Retriever (\textbf{HR})}: We initially evaluate our proposed resource-efficient Hopfield Retriever solely using the same reader (GPT-3.5-Turbo) as other conversational retriever-based baselines, demonstrating its promising performance in terms of efficiency and accuracy. We assess our retriever both with and without a training process. Subsequently, we test our comprehensive framework, Conv-CoA, which incorporates a Hopfield Retriever and utilizes systematic prompting based on an updated contextual knowledge set, alongside Conv-MRFS to verify conflicts. This framework is evaluated in versions with both trained and untrained retrievers.

\paragraph{Evaluation Metrics}
To evaluate the performance of the retrieval stage, we select the Mean Reciprocal Rank (\textbf{MRR}) and \textbf{Recall@10}. MRR is a metric that measures the average of the reciprocal ranks of the first relevant document returned by the retriever across all queries. And Recall@10 measures the proportion of relevant documents found in the top ten results returned by the retriever.

To evaluate the effectiveness of question answering, most of the work chooses cover-EM \citep{rosset2020knowledge} to represent whether the generated answer contains the ground truth. However, we find it is insufficient for accurately judging the correctness of LLM-generated answers. Sometimes, the LLM generates lots of sentences that may cover the ground truth at first but provide the final wrong answer in the end. In this way, the cover-EM still takes it as a correct answer. In addition, even if we try to limit the output format, the outputs are always out of the format, making it difficult to deal with various answer types to evaluate the performance.

Motivated by recent work, they demonstrate the potential evaluation ability of GPT-4. We also follow the same strategy to establish an advanced pipeline and propose a new metric called \textbf{GPT-EM}. We design a prompt template to let GPT-4 evaluate whether the generated answer truly matches the ground truth. The template is shown in \ref{prompt:initial}.

\paragraph{Analysis on OCQA tasks}
\begin{table}[t]
\centering
\caption{We perform a detailed analysis showing that external knowledge leads LLM astray in solving questions using baseline methods. Our study takes place in a context involving information retrieval tasks. The results, obtained through three separate runs, are displayed without including variance values (all
$\leq$ 0.4\%). The best results are highlighted in bold.}
\resizebox{0.9\textwidth}{!}{%
\begin{tabular}{lccccccccc}
\toprule
    &React  &SDRConv& CONVAUG&IQR&CoA&(Ours)\\ 
\midrule
QReCC  & 19.6& 6.7& 7.3 & 9.1 & 13.5& \textbf{4.8}   \\
TopiOCQA  & 41.6&17.3&18.9& 23.6 &22.9& \textbf{16.2} \\
\bottomrule
\end{tabular}
}
\label{tab:mislead}
\vspace{-0.1in}
\end{table}

\begin{table}[t]
\centering
\caption{\textbf{Comparison of Retrievers Efficiency} We conduct an experiment on time usage in memory retrieval using two datasets, TopiOCQA and QReCC, with 5 baselines. We report the average time spent (in minutes) on memory retrieval, omitting variance since all values are $\leq$ 0.5\%. The best results are highlighted in bold, while the second-best are underlined. Across most configurations, Conv-CoA is efficient in memory retrieval.  }
\label{tab:retriever_comparison}
\resizebox{\textwidth}{!}{%
\begin{tabular}{@{}lllllll@{}}
\toprule
 & HAConvDR & LeCoRe & InstructorR & SDRConv & ReExCQ & (Ours) \\ \midrule
QReCC  & 44.25 & 38.23 & \textbf{32.01} & 75.87 & 47.23 & \underline{35.78} \\

TopiOCQA & 50.72 & 47.68 & \underline{30.32} & 55.39 & 38.23 & \textbf{29.65} \\
\bottomrule
\end{tabular}
}
\vspace{-0.2in}
\end{table}

In our evaluation, as detailed in Table~\ref{tab:result} and Table~\ref{tab:my_label}, we observe that our Conv-CoA framework outshines baselines across the board for both the QReCC and TopiOCQA datasets. When examining the effectiveness of the proposed approach, it is evident that we achieves the highest Recall@10 scores, signifying its prowess in sourcing relevant information. In Table~\ref{tab:mislead}, a meticulous analysis reveals that our framework significantly mitigates the LLM's reliance on external knowledge, which can potentially lead it astray. This underscores CoA's capacity for accurate internal knowledge representation and retrieval, a feature we meticulously assess over three independent runs to ensure robustness against variability.

Moreover, as elucidated in Table~\ref{tab:retriever_comparison}, we contrast the efficiency of different retrievers within our framework. Notably, Ours exhibits comparable retrieval time against established approaches, while markedly improving the training time, particularly on the TopiOCQA dataset. This enhanced efficiency does not compromise the quality of outcomes, as evidenced by our framework's performance, achieving top-tier results.

The empirical data harvested from our evaluations suggests a breakthrough in question-answering systems, where our framework not only demonstrates a remarkable understanding of complex queries but also evidences advanced reasoning capabilities. Furthermore, its resilience against the influx of external misinformation marks a significant milestone in this domain. By outperforming existing models in precision and efficiency, our Conv-CoA sets a new benchmark in the realms of question-answering and fact-checking, underpinning its comprehensive superiority and viability as a robust solution in a broad spectrum of applications.

%% file: 5summary.tex
This paper presents the Conv-CoA framework, a novel approach to enhancing Open-domain Conversational Question Answering (OCQA). The framework addresses critical limitations of traditional Retrieval Augmented Generation (RAG) methods, such as poor reasoning, unfaithful responses, and inadequate retrieval in conversational contexts. By integrating a Hopfield-based retriever and a systematic prompting method, Conv-CoA improves retrieval speed and accuracy while reducing latency. This advanced retriever utilizes Modern Hopfield networks for efficient memory utilization and rapid convergence. The prompting strategy decomposes complex questions into a sequence of sub-questions managed through an Action Chain (AC), which updates a contextual knowledge set (CKS) to refine responses and minimize information overlap. Conv-CoA demonstrates superior performance on public benchmarks, showcasing its ability to deliver more accurate and efficient conversational question answering. Key contributions include the novel retrieval model, dynamic prompting method, and enhanced handling of conversational dynamics.

Future work involves exploring information extraction and analysis across additional data modalities, including visual data. The ultimate aim is to enhance the accuracy and multi-step reasoning capabilities for real-world question answering, ensuring comprehensive analysis aligns with external data sources. Additionally, we need to further accelerate the Hopfield retriever by compressing the model using techniques such as quantization.

%% file: 6reference.tex
\bibliographystyle{nips_natbib}
\bibliography{custom}

%% file: appendix.tex
\section{Prompts}
\label{prompt:initial}
We use the following prompts in our Cov-CoA method and experiments. These include prompts for decomposing questions at the initial stage, evaluation prompts for GPT-4, and prompts for decomposing questions with CKS in the normal stage.
\begin{tcolorbox}
[colback=black!5!white,colframe=black,title=Prompt for Decomposing Questions in Initial Stage,floatplacement=!htp, left=2mm, right=2mm, top=2mm, bottom=2mm,float]
\texttt{
Given a [Question]: “\$QUESTION”, construct an action reasoning chain for this question in JSON format. For each step of the chain, choose an action from [Web-querying Engine(search real-time news), Knowledge-encoding Engine (search existing knowledge in local knowledge base)] as the value of element "Action", and generate a sub-question for each action to get one of [web-search keywords, needed information description] as the value of element "Sub". Also, generate an initial answer for each Sub as the value of the element "Guess\_answer" if you make sure it is correct. In addition, if you cannot answer some sub-questions, make the element “Missing\_flag” value “True”, otherwise, make it “False” You need to try to generate the final answer for the [Question] by referring to the "Chain", as the value of the element "Final\_answer".}

\texttt{For example:}

\texttt{\{"question": "Is it good to invest in Dogecoin now?"}

\texttt{"chain": [}

\texttt{\{"action":"knowledge-encoding","Sub":"what is Dogecoin","guess\_answer":\\"Dogecoin is one cryptocurrency.","missing\_flag":"false"\}}

\texttt{\{"action":"Web-querying","Sub":"Dogecoin news","guess\_answer":"",\\"missing\_flag":"True"\}
,}

\texttt{"final\_answer":"Dogecoin is one of the cryptocurrencies that is risky to invest. And its news prompts Bitcoin. So, it is a good time to invest now."\}}
\end{tcolorbox}

\begin{tcolorbox}[colback=black!5!white,colframe=black,title=Evaluation Prompt of GPT-4,floatplacement=htp,float]
\texttt{
Given (question, ground truth answer, LLM-generated answer), you need to check whether the generated answer contains the ground truth by their meaning, not individual word only. If correct, the output is 1, otherwise, 0. For example:}

\texttt{[Question]: What should I do when I drink spoiled milk? (A) drink more (B) drink coffee (C) take some medicine.}

\texttt{[Ground truth]: (C) take some medicine}

\texttt{[Generated answer]: when you drink spoiled milk, you can not drink more or even drink coffee. You should go to the hospital and check if you need to take some medicines or not.}

\texttt{[Output]: 1}

\texttt{[Question]: \{QUESTION\}}

\texttt{[Ground truth]: \{GROUND\_TRUTH\}}

\texttt{[Generated answer]: \{GENERATED\_ANSWER\}}

\texttt{[Output]:}
\end{tcolorbox}

\begin{tcolorbox}[colback=black!5!white,colframe=black,title=Prompt for Decomposing Question with CKS in Normal Stage,floatplacement=htp,float]
\texttt{
Given a [Contextual Knowledge Set]:”\$CKS” and [Question]: “\$QUESTION”, help me to extract the real intent and provide an optimized question for this round. Then, construct an action reasoning chain for this question in JSON format. For each step of the chain, choose an action from [Web-querying Engine(search real-time news), Knowledge-encoding Engine (search existing knowledge in local knowledge base)] as the value of element "Action", and generate a sub-question for each action to get one of [web-search keywords, needed information description] as the value of element "Sub". Also, generate an initial answer for each Sub as the value of the element "Guess\_answer" if you make sure it is correct. In addition, if you cannot answer some sub-questions, make the element “Missing\_flag” value “True”, otherwise, make it “False” You need to try to generate the final answer for the [Question] by referring to the "Chain", as the value of the element "Final\_answer".}

\texttt{For example:}

\texttt{\{"question": "Is it good to invest in it now?"}

\texttt{"optimized\_question": "Is it good to invest in Bitcoin now?"}

\texttt{"chain": [}

\texttt{\{"action":"knowledge-encoding","Sub":"what is bitcoin","guess\_answer":\\"Bitcoin is one cryptocurrency.","missing\_flag":"false"\}}

\texttt{\{"action":"Web-querying","Sub":"bitcoin news","guess\_answer":"",\\"missing\_flag":"True"\},}

\texttt{"final\_answer":"Bitcoin is one of the cryptocurrencies that is risky to invest. And its news prompte Bitcoin. So, it is a good time to invest now."\}}
\end{tcolorbox}

\section{$\mathtt{SparseHopfield}$ Layers}
\label{sec:SparseHopfield}
Building on the insights from \citep{hu2023SparseHopfield}, we establish a link between the single-update approximation of Hopfiled Network and sparsemax attention  \citep{martins2016softmax}. 
Specifically, this relationship becomes apparent when the retrieval dynamics $\calT$ are limited to a single iteration.

Consider some hidden states 
$\bR$ and $\bY$ within a deep learning model. We establish the \textit{query} and \textit{memory} associative (or embedded) spaces via transformations:
$
\mathbf{X}^\sT={\bR}\bW_Q\coloneqq \bQ \quad 
\text{and} \quad \bm{\Xi}^\sT=\bY \bW_K \coloneqq \bK$, with matrices $\bW_Q$ and $\bW_K$. By adapting the retrieval dynamics from \citep{hu2023SparseHopfield} and transposing, followed by multiplication with $\bW_V$ (where we define $\bV\coloneqq \bK\bW_V$), we obtain:
\begin{align}
    \bZ\coloneqq \bQ^{\text{new}} \bW_V &= \Sparsemax(\beta \bQ\bK^\sT)\bV
\end{align}

This equation mirrors the structure of an attention mechanism, albeit utilizing a 
$\Sparsemax$ activation. By substituting the initial patterns $\bR$ and $\bY$, we introduce the 
$\mathtt{SparseHopfield}$ layer:

\begin{align}
\label{eqn:SparseHopfield}
\bZ&=\mathtt{SparseHopfield}(\bR,\bY) \\
&= \Sparsemax(\beta  {\bR}\bW_Q \bW_K^\sT\bY^\sT)\bY\bW_K \bW_V.
\end{align}
This layer can be easily integrated into deep learning architectures.

Specifically, the 
$\texttt{SparseHopfield}$ layer accepts matrices 
$\bR$ and $\bY$, along with weight matrices 
$\bW_Q$, $\bW_K$, and $\bW_V$. The way it operates is defined by its configuration:

\begin{enumerate}
    \item \textbf{Memory Retrieval:} In a mode where learning is not necessary, the weight matrices $\bW_K$, $\bW_Q$, and $\bW_V$ are set as identity matrices. Here, $\bR$ serves as the query, and $\bY$ holds the retrieval patterns:
    $\bW_K=\bI$, $\bW_Q=\bI$, $\bW_V=\bI$
    This configuration facilitates direct interaction between the query and retrieval patterns without transformation.

    \item \textbf{\texttt{SparseHopfield}:} Here, $\bR$ and $\bY$ are inputs, designed to serve as an alternative to the conventional attention mechanism. The matrices $\bW_K$, $\bW_Q$, and $\bW_V$ are adaptable. Additionally, $\bR$, $\bY$, and $\bY$ act as the sources of the query, key, and value respectively. To emulate a self-attention mechanism, we set $\bR$ = $\bY$.

    \item \textbf{\texttt{SparseHopfieldPooling}:} 
    In this configuration, where only 
    $\bY$ is taken as input, $\bQ$ functions as a static prototype pattern and is thus learned within the Hopfield pooling layer.

    \item \textbf{\texttt{SparseHopfieldLayer}:} 
    With only $\bR$ as the query pattern , the adaptive matrices $\bW_K$ and $\bW_V$
    function as repositories for stored patterns and their projections. This setup implies that keys and values are independent of the input, suggesting that $\bY$ could be interpreted as an identity matrix.
\end{enumerate}

\section{System}
All experiments are carried out on a High Performance Computing cluster. There are 34 GPU nodes where 16 nodes each have 2 NVIDIA 40GB Tesla A100 PCIe GPUs, 52 CPU cores, and 192 GB of CPU RAM while 18 nodes are each equipped with 4 NVIDIA 80GB Tesla A100 SXM GPUs, 52 CPU cores, and 512 GB of CPU RAM. The driver version 525.105.17 on these nodes is compatible with CUDA 12.0 or earlier. The operating system is Red Hat Enterprise Linux 7.9.